\documentclass[letterpaper]{article} % DO NOT CHANGE THIS
\usepackage[preprint]{aaai2027}  % DO NOT CHANGE THIS

\usepackage[hyphens]{url}  % DO NOT CHANGE THIS
\usepackage{graphicx} % DO NOT CHANGE THIS
\usepackage{natbib}  % DO NOT CHANGE THIS AND DO NOT ADD ANY OPTIONS TO IT
\usepackage{caption} % DO NOT CHANGE THIS AND DO NOT ADD ANY OPTIONS TO IT
\usepackage{algorithm}
\usepackage{algpseudocode}

\usepackage{newfloat}
\usepackage{listings}
\DeclareCaptionStyle{ruled}{labelfont=normalfont,labelsep=colon,strut=off} % DO NOT CHANGE THIS
\floatstyle{ruled}
\newfloat{listing}{tb}{lst}{}
\floatname{listing}{Listing}

\usepackage{booktabs}

\usepackage{times}
\usepackage{latexsym}
\usepackage{booktabs}
\usepackage[T1]{fontenc}
\usepackage{amssymb}
\usepackage{multirow}
\usepackage{tabularx}
\usepackage{amsfonts}

\usepackage[utf8]{inputenc}
\usepackage{microtype}
\usepackage{inconsolata}
\usepackage{xspace}
\usepackage[inline]{enumitem}

\usepackage{url}

\usepackage{booktabs}
\usepackage{amsfonts}
\usepackage{amssymb}
\usepackage{amsmath}
\usepackage{nicefrac}
\usepackage{xcolor}
\usepackage{algpseudocode}
\usepackage{siunitx}
\usepackage{tcolorbox}
\usepackage{colortbl}
\definecolor{bcsblue}{HTML}{DCEBFA}
\definecolor{hmgreen}{HTML}{DCF0DD}

\title{How Context Attribution Handles What the Model Already Knows}
\author{
    Quoc-Huy Trinh\textsuperscript{\rm 1}\equalcontrib\thanks{Corresponding authors.},
    Lin Zhu\textsuperscript{\rm 1}\equalcontrib,
    Sebastian Szyller\textsuperscript{\rm 1}\footnotemark[2]
}
\affiliations{
    \textsuperscript{\rm 1}Aalto University\\
    \{quoc.h.trinh, lin1.zhu\}@aalto.fi, contact@sebszyller.com
}

\DeclareMathAlphabet{\mathsfit}{\encodingdefault}{\sfdefault}{m}{sl}
\SetMathAlphabet{\mathsfit}{bold}{\encodingdefault}{\sfdefault}{bx}{n}

\newcommand{\dtrain}{\ensuremath{D_\mathrm{train}}\xspace}

\newcommand{\dforget}{\ensuremath{D_\mathrm{forget}}\xspace}

\newcommand{\wmdppp}{WMDP-Cyber++\xspace}

\newcommand{\attr}{\ensuremath{\mathrm{Attr}}\xspace}

\newcommand{\interface}{\mathcal{I}}

\newcommand{\model}{\ensuremath{p_{\theta}}\xspace}
\newcommand{\aps}{\ensuremath{\mathrm{APS}}\xspace}
\newcommand{\util}{\ensuremath{\mathrm{Util}}\xspace}
\newcommand{\rmi}{\ensuremath{\mathrm{RmI}}\xspace}
\newcommand{\ssp}{\ensuremath{\mathrm{SSP}}\xspace}
\newcommand{\rbench}{\ensuremath{R_{\mathrm{bench}}}\xspace}

\usepackage{acro}
\DeclareAcronym{ml}{short   = ML,   long = machine learning}
\DeclareAcronym{llm}{short  = LLM,  long = large language model}

\DeclareAcronym{icl}{short  = ICL,  long = in-context learning}
\DeclareAcronym{iw}{short   = IW,   long = in-weight}

\DeclareAcronym{loo}{short  = LOO,  long = leave-one-out}

\DeclareAcronym{aps}{short  = APS,  long = attribution preservation score}
\DeclareAcronym{bcs}{short  = BCS,  long = base-model context attribution score}
\DeclareAcronym{cac}{short  = CAC,  long = cross-model context attribution consistency}
\DeclareAcronym{lds}{short  = LDS,  long = linear data-modeling score}
\DeclareAcronym{ssp}{short  = SSP,  long = source separation precision}
\DeclareAcronym{rbo}{short  = RBO,  long = rank-biased overlap}

\begin{document}

\maketitle
\begin{abstract}
Context attribution methods for \acp{llm} identify
which input context contributes to the model response.
Recent works show the initial success in attributing the contributive score of the contexts.
However, we observe that
when the context overlaps with the training data, 
these methods cannot disentangle in-context from \ac{iw} contributions,
producing unreliable scores.
Based on this observation, in this work, we introduce:
\begin{enumerate*}[label=\arabic*)]
    \item an evaluation protocol that relies on four new metrics (\ac{bcs}, \ac{cac}, \ac{aps}, \ac{ssp}) and 
    \item a benchmark dataset (\wmdppp) with ground-truth provenance labels
\end{enumerate*}
to systematically assess attribution under \ac{iw} overlap.
In our experiments across four well-known context attribution methods,
we demonstrate that they provide unfaithful attribution when the knowledge from the context also exists in the weights.
Finally, we adapt these methods for source separation (\ac{iw} vs. \ac{icl})
and show that they cannot do the disentanglement based on the contributive score.

% and the distortion is uncorrelated with memorization strength, 
% ruling out post-hoc calibration, 
% suggesting that better attribution methods should be proposed to handle the context in the training data, 
% but also do the context source disentanglement to release the faithful and reliable results.
\end{abstract}
\acresetall
\section{Introduction}
Context attribution in \acp{llm} is the task of identifying which part of the context contributes to the model's response \cite{cohen-wang2024contextciteattributingmodel}.
It enables users to refine the context for better responses based on the attribution scores and identify unexpected prompt-response dependencies (e.g. prompt injection)~\cite{wangtracllmgenericframework}. 
The underlying question
-- what it means for a segment to \emph{contribute} -- 
can be read at two levels:
\begin{enumerate*}[label=\arabic*)]
\item whether the \emph{text} of a segment contributes to the response.
\item whether the \emph{knowledge} it carries contributes to the response.
\end{enumerate*}

Recent context attribution methods rely on different mechanisms
(e.g., leave-one-out and Shapley-based methods)
to estimate which segments of the input context contribute the most to the model output.
In this work, 
we consider ContextCite~\cite{cohen-wang2024contextciteattributingmodel}, AttriBot~\cite{liu2024attribotbagtricks}, 
TracLLM~\cite{wangtracllmgenericframework} and TokenShapley~\cite{xiao2025tokenshapleytokenlevel,khandelwal2019generalizationmemorizationnearest}.
% sparse surrogate modeling over perturbed contexts~\cite{cohen-wang2024contextciteattributingmodel},
% leave-one-out-style attribution~\cite{liu2025attribotbagtricks},
% perturbation-based trace-back for long-context settings~\cite{wangtracllmgenericframework},
% and Shapley-style token-level attribution with retrieval-based estimation~\cite{xiao2025tokenshapleytokenlevel,khandelwal2019generalizationmemorizationnearest}.
\begin{figure}[!t]
    \centering
    \includegraphics[width=0.47\textwidth]{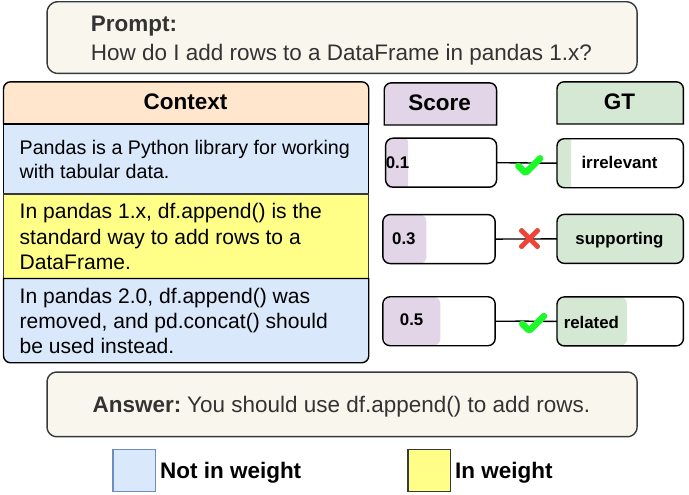}
    \caption{Current context attribution methods cannot distinguish
    \acs{iw} overlap from true irrelevance. 
    The yellow context supports the LLM's response but its attribution score is lower because the same information is also drawn from the weights.}
    \label{fig:problem_statement}
\end{figure}
These methods primarily measure importance through the sensitivity of the model's response to perturbations of the input context, and operate exclusively at the first level.
Such scores are easy to interpret when the context segment is the sole carrier of the suprooting knowledge,
and there is no knowledge in the weights,
where two levels are equivalent.
However, once \ac{icl} and \ac{iw} knowledge overlap, the levels diverge:
a supporting segment, that also exists in the weights, may receive a near-zero score because removing it does not impact the output. 
A related symptom has been observed in prior work~\cite{cohen-wang2024contextciteattributingmodel},
where a source supporting the answer may receive a low score.

In real-world applications, 
this misalignment can make a genuinely supporting document span appear unimportant, 
leading systems to discard evidence that is critical for verifying the response.
Figure~\ref{fig:problem_statement} illustrates this issue with a version-sensitive documentation example.
The supporting segment correctly states that \texttt{df.append()} can be used in pandas 1.x, 
but the model may already know this from its weights.
Thus, existing context attribution methods may assign the segment a low score even though it provides task-relevant support.

This exposes two limitations:
\begin{enumerate*}[label=\arabic*)]
\item the attribution score is difficult to interpret.
A low score means the segment is irrelevant or
that its knowledge is already in the weights;

\item current evaluation metrics cannot expose this failure, e.g., 
top-$k$ log-probability drop and \acs{lds} 
\cite{cohen-wang2024contextciteattributingmodel} are themselves removal-based.
Thus, a method can distort under \ac{iw} overlap while still ranking well.
\end{enumerate*}

Motivated by these gaps, we ask:

\begin{tcolorbox}[colback=gray!5, colframe=black, boxrule=0.5pt, arc=0pt]
\centering
How do context attribution methods handle 

what the \ac{llm} already knows?
\end{tcolorbox}

In this work, 
we answer this by controlling the \ac{iw} knowledge directly. 
% Starting from a base model, 
% we finetune it on a known corpus, 
% yielding a pair of models that differ only in whether that knowledge is in the weights.
% Holding the context fixed across the pair isolates how attribution responds to what the model already knows, 
% evaluated by task-level correctness metrics rather than likelihood-based metrics. 
% In summary, we make three contributions:
We make the following contributions:
\begin{itemize}
    \item We introduce an evaluation protocol based on four new metrics
    (\acs{bcs}, \acs{cac}, \acs{aps}, \acs{ssp}).
    The protocol captures how the attribution scores change when the same knowledge is available in both the context and the weights.

    \item We release \emph{WMDP-Cyber++}, a new dataset that provides
    ground-truth source attribution (\ac{iw} vs.\ \ac{icl}) for each context segment, 
    addressing the absence of benchmarks for \ac{iw}/\ac{icl} source attribution.

    \item We demonstrate that evaluated attribution methods assign varying scores
    depending on whether the information is provided only in the context or also in the weights. Additionally, we establish that these methods cannot be simply adapted for source separation.
\end{itemize}
\section{Background}

\paragraph{Large language models.} 
\acp{llm} are commonly formulated as autoregressive conditional probability models.
A prompt $X$ which corresponds to the questions or instructions is concatenated with a set of context units $C=\{c_1,c_2,\ldots,c_k\}$.
The model \model then generates a response sequence 
$Y=(y_1,y_2,\ldots,y_t)$ as

\begin{equation}
\model(Y \mid C \oplus X)
=
\prod_{t=1}^{T} p_\theta(y_t \mid C \oplus X, Y_{<t}),
\end{equation}
where $Y_{<t}=(y_1,\ldots,y_{t-1})$ and $\oplus$ denotes sequence concatenation.
The context $C$ may include retrieved documents, 
in-context examples, dialogue history, or other external information.

\paragraph{Context attribution.}
Context attribution aims to attribute a generated response back to specific parts of the context.
Given a prompt $X$, 
a context set $C=\{c_1,\ldots,c_k\}$, 
and a response $Y$ generated by \model\ conditioned on $C \oplus X$,
an attribution method $\attr(\cdot)$ quantifies how each unit $c_i$ contributes to $Y$ and assigns an importance score
$s_i \in \mathbb{R}$ to each context unit $c_i$, 
yielding the attribution score set
\begin{equation}
s_i = \attr(c_i, Y, C, X),
\end{equation}
a larger value of $s_i$ indicates that $c_i$ makes a stronger contribution to $Y$.

While numerous attribution methods have been proposed, 
their quality must also be evaluated. 
Existing evaluation metrics assess whether attribution scores faithfully reflect context contributions to the original response. 
\emph{Top-$k$ log-probability drop} evaluates whether the highest-ranked context units are the most contributive, 
as removing them should maximally reduce the likelihood. 
Furthermore, \emph{\ac{lds}} evaluates whether attribution scores predict the relative effects of different context ablations.
However, they remain likelihood-based and do not directly evaluate task-level correctness.

\section{Problem statement}
Given an input prompt $X$, 
a set of context examples $C = \{c_1, c_2, \ldots, c_k\}$, 
and a generated response $Y$ produced by an \ac{llm} under the \ac{icl} paradigm,
the goal of any \attr is to attribute the contributive score of the individual context elements in $C$ that lead to the generation of $Y$.
In this work, 
we evaluate attribution at the knowledge level rather than only at the text level. 
Therefore, we define attribution quality using task-level correctness:
a context segment or subset is important if it supports the model in answering the correct answer, 
not merely if it changes the surface form or likelihood of $Y$.
Under this view,
the context unit $c$ is \emph{necessary} when removing it from the full context
$C$ leads to the incorrect response $Y$.
It is \emph{sufficient} when presenting $c_i$ within a small
top-$k$ subset is enough to recover $Y$ without the remaining context.

% Additionally, $c$ is \emph{necessary} when ....
% and it is \emph{sufficient} when ...

A common question in context attribution is whether $c_i$ that shapes the model's output originates from \ac{icl} or instead reflects knowledge from the weights (\ac{iw} knowledge). 
Therefore, in this work, we aim to disentangle $c_i \in$ \dtrain from $c_i \notin$ \dtrain, for some training set \dtrain.
We identify two criteria that a successful attribution method should satisfy:
\begin{enumerate*}[label=\textbf{C\arabic*}]
    \item\label{cr-fidelity} (attribution fidelity)
    \attr assigns a high $s_i$ to $c_i$ if it contributes to a correct response, whether it is explicitly included \ac{icl} or implicitly \ac{iw};
    \item\label{cr-disentanglement} (data-source disentanglement evaluation)
    any \attr must be able to distinguish whether the influence of $\{c_1, c_2, \ldots, c_k\} \in C$ originates from \ac{iw} or \ac{icl}.
    % \item\label{cr-corr-necessity} (task-level correctness and necessity)
\end{enumerate*}

These criteria require a more comprehensive evaluation protocol for context attribution.
Rather than relying only on likelihood-based changes, 
such a protocol should also use task-level correctness to evaluate whether attributed segments are sufficient and necessary for answering correctly.
It should also include a way to test whether attribution methods can disentangle support from \ac{icl} and \ac{iw} knowledge.
% Therefore, there must exist an evaluation protocol that reflects the success/failure of \attr at disentanglement.
\section{Proposed evaluation protocol}
\label{sec:evaluation_benchmark}

% However, they do not directly test whether the selected context units are sufficient to support task-level correctness,
% nor whether they are necessary under the full-context condition.
To address criteria~\ref{cr-fidelity},~\ref{cr-disentanglement},
we introduce an evaluation protocol built around four new metrics.
Unlike prior evaluation that relies on likelihood-based metrics, 
our protocol grounds attribution in task-level correctness. 
We argue that task-level correctness provides a more direct measure of knowledge presence, 
as it captures the model's ability to apply knowledge to solve a task, 
whereas likelihood-based metrics merely assess the model's confidence in token
prediction over a corpus.

\subsection{Base-model context attribution score}
We first introduce \ac{bcs} that measures the sufficiency and necessity of the top-$k$ subset.
\ac{bcs} is computed on the base model which has not been exposed to \dtrain,
thus, 
the attribution directly reflects the model's reliance on the provided context rather than on \ac{iw} knowledge.

For each $X$,
we evaluate four conditions that differ exclusively in the provided context:
\begin{enumerate*}[label=\arabic*)]
    \item \texttt{no-ctx} (no context),
    \item \texttt{full-ctx} (full context),
    \item \texttt{rm-topk} (full context with top-$k$ removed),
    \item \texttt{topk-only} (only top-$k$ retained).
\end{enumerate*}
Let $z \in \mathcal{Z} = \{\texttt{no-ctx}, \texttt{full-ctx}, \texttt{rm-topk}, \texttt{topk-only}\}$ index the condition, $\mathrm{Acc}_{z}^{\model}$ denote the task accuracy of \model under condition $z$,
and define the full-context gain as:
\begin{equation}
\Delta^{\model} = \mathrm{Acc}_{\text{\texttt{full-ctx}}}^{\model} - \mathrm{Acc}_{\text{\texttt{no-ctx}}}^{\model} .
\label{eq:delta_g}
\end{equation}
\ac{bcs} establishes the attribution quality of the base model.
Two complementary properties are required for a top-$k$ selection to faithfully capture the evidence the model relies on.

\noindent\textbf{Top-$k$ utility} (\util) measures \emph{sufficiency} -- how much performance the base model recovers using only top-$k$:
\begin{equation}
    \util =
    \frac{
        \mathrm{Acc}_{\text{topk-only}}^{p_{\theta_\text{base}}} - \mathrm{Acc}_{\text{no-ctx}}^{p_{\theta_\text{base}}}
    }
    {
    \Delta^{p_{\theta-\text{base}}}
    } .
\end{equation}

\noindent\textbf{Top-$k$ removal impact} (\rmi) measures \emph{necessity}---how much the base model loses when top-$k$ is removed:
\begin{equation}
    \rmi =
    \frac{
    \mathrm{Acc}_{\text{full-ctx}}^{p_{\theta_\text{base}}} - \mathrm{Acc}_{\text{rm-topk}}^{p_{\theta_\text{base}}}
    }
    {
    \Delta^{p_{\theta-\text{base}}}
    } .
\end{equation}
We combine the two via a geometric mean:
\begin{equation}
    \mathrm{BCS} = \sqrt{\util \cdot \rmi} .
\label{eq:score_c}
\end{equation}
Either property alone is insufficient.
High \util with low \rmi indicates that the top-$k$ segments are sufficient on
their own, 
yet removing them does not hurt performance, 
because the remaining context segments can supply the same information.
High \rmi with low \util indicates the opposite: 
the top-$k$ segments are necessary, 
since removing them degrades performance, 
yet they are not sufficient on their own, 
as answering correctly also requires other segments in the context.
% and thus indicates that top-$k$ has been empirically verified as the genuine in-context evidence carrier for the base model.

\subsection{Cross-model context attribution consistency}

While \ac{bcs} evaluates whether the selected top-$k$ context units are sufficient and necessary for task-level correctness on the base model,
it does not assess whether the attribution ranking remains consistent after fine-tuning.
We therefore introduce \ac{cac} to compare the attribution rankings produced on the base model and the fine-tuned model.

For each $X$,
we sort context units by attribution score on $p_{\theta_{\text{base}}}$ and $p_{\theta_{\text{ft}}}$,
yielding ranked lists $\pi^{p_{\theta_{\text{base}}}}$ and $\pi^{p_{\theta_{\text{ft}}}}$.
We measure their agreement using \ac{rbo},
which gives higher weight to agreement at the top ranks:
\begin{equation}
    \mathrm{CAC} = \mathrm{RBO}\!\left(\pi^{p_{\theta_{\text{base}}}}, \pi^{p_{\theta_{\text{ft}}}}; \rho\right),
\end{equation}
where $\rho \in (0,1)$ controls the top-weight decay.

A faithful method should remain consistent: 
a segment that contributes to the answer still contributes whether or not the same knowledge is present as \ac{iw} knowledge, 
so its attribution should not change. 
A low \ac{cac} therefore indicates that a method's scores are driven by the model's knowledge state rather than by each segment's contribution.

\subsection{Attribution preservation score}
While a high \ac{cac} shows that a method's ranking is stable across the two
models, 
stability alone is insufficient: 
a method could obtain a high \ac{cac} by producing consistently incorrect rankings on both models.
We therefore combine \ac{bcs} and \ac{cac} into a single metric, \ac{aps}, as their harmonic mean:
\begin{equation}
    \mathrm{APS}
    =
    \frac{2 \cdot \mathrm{BCS} \cdot \mathrm{CAC}}
    {\mathrm{BCS} + \mathrm{CAC}}.
    \label{eq:aps}
\end{equation}
Achieving a high \ac{aps} requires strong performance across both \ac{bcs} and \ac{cac}.

\subsection{In-context/in-weight source separation precision}

\ac{bcs}, \ac{cac}, and \ac{aps} assess whether the attributed context is necessary, sufficient and the attribution scores are consistent across models.
None of them test whether \attr assigns each segment to the correct evidence source.
Since our evaluation requires source-specific predictions,
we introduce source separation precision (\ssp), which measures whether \attr correctly identifies the source as \ac{iw}/\ac{icl}.

Measuring this requires retaining only the samples that the fine-tuned model answers correctly without the context, while the base model fails.
This indicates that the model learned the information.

\begin{equation}
\rbench
=
\{i: p_{\theta_{\mathrm{ft}}}(X_i)=A_i
\,\wedge\,
p_{\theta_{\mathrm{base}}}(X_i)\neq A_i\},
\end{equation}
where $X_i$ is the question-only prompt and $A_i$ is the gold answer, both drawn from \dtrain.

For each sample in $\rbench$, let $k_i$ be the number of context
segments for the sample $X_i$;
\attr is applied to $p_{\theta_{\mathrm{ft}}}$ to produce a contributive score $s_i$ for each context segment $c_j$.
Since existing attribution methods output scalar contribution scores rather than explicit source labels, 
we introduce the disentanglement interface $\interface$ that adapts them to also output a source label \ac{iw}/\ac{icl}.
Specifically, for sample $i$ and segment $j$, the interface outputs
$\interface^{pred}_{ij} \in \{\text{IW}, \text{ICL}\}$
(we describe the detailed process in Section~\ref{sec:experiment}).

\ssp measures the proportion of context segments whose predicted source
matches their controlled source label:
% \begin{equation}
% \ssp = \frac{\sum_{i < |\rbench|}|\{o^{pred}_{i} = o^{gt}_i\}|}{|\rbench|}
% \label{equa:ssp}
% \end{equation}

\begin{equation}
    \text{SSP} = \frac{\sum_{i \in \rbench} \sum_{j=1}^{k_i}
1\!\left[\, \interface_{ij}^{pred} = \interface_{ij}^{gt} \,\right]}
    {\sum_{i \in \rbench} k_i}
    \label{equa:ssp}
\end{equation}

A high \ssp indicates that \attr can correctly distinguish whether each context
segment contributes through \ac{icl} evidence or through \ac{iw} knowledge.
Since WMDP-Cyber++ contains a balanced number of \ac{iw} and \ac{icl} segments,
random source assignment achieves an expected \ssp of $0.5$.
It is a controlled diagnostic for methods that explicitly emit in-context and \ac{iw} attribution labels.

\subsection{WMDP-Cyber++ dataset}
\label{subsec:dataset}
Based on the cyber subset of the WMDP benchmark~\cite{doshi2025doesunlearningtruly},
we construct \wmdppp,
an augmented benchmark with source-labeled context segments.
Each data point is a multiple-choice cybersecurity question with a mixed-source context comprising segments whose provenance (\ac{iw} or \ac{icl}) is known by construction
\footnote{Prompts for each step are provided in the appendix.}.

\paragraph{Benchmark construction.}
Let $\dtrain$ denote the \texttt{cyber-retain-corpus} used to finetune the target model,
and let $\dforget$ denote the held-out \texttt{cyber-forget-corpus} that is \emph{never} seen during training.
For each WMDP-Cyber question $X_i$ with ground-truth answer $A_i$,
we construct a context $C_i = [c_{i,1}, c_{i,2}, \ldots, c_{i,k}]$ consisting of $k$ segments,
where $k = k_{\text{iw}} + k_{\text{icl}}$:
$k_{\text{iw}}$ is number of \ac{iw} segments retrieved from \dtrain,
and $k_{\text{icl}}$ is number of in-context segments synthesized from \dforget.
The segments are shuffled so that their ordering carries no provenance signal.
% In our experiments, we set $K_{\text{iw}} = K_{\text{icl}} = 3$.

\paragraph{Step 1: \ac{iw} segment retrieval.}
We chunk documents in $\dtrain$ into 500-token passages with 50-token overlap at the boundaries.
For retrieval, we index them using the Qwen Embedding model~\cite{zhang2025qwen3embeddingadvancing}.
For each question, 
we form a query by concatenating the full question $X_i$ with the gold answer $A_i$,
retrieve the top-$k_{\text{cand}}$ candidate passages,
and re-rank them with GPT-4o~\cite{openai2024gpt4osystemcard}.
To re-rank, the model is instructed to score each passage on a 10-point scale based on its relevance to the concepts and knowledge required to answer the question.
We retain the top-$k_{\text{iw}}$ passages as $S_i^{\text{iw}}$ and label them as \ac{iw}.
This corresponds to Algorithm~\ref{algo:segment_retrieval} with
$z=\text{iw}$, $k_{\text{cand}}=10$, and $k_{\text{iw}}=3$.

% list of size 10 <- retrieve
% list of size 10 reranked <- rerank (list of size 10)
% list of size 10 reranked labelled <- label as iw

\begin{algorithm}[t]
\caption{IW/ICL segment retrieval}
\label{algo:segment_retrieval}
\begin{algorithmic}[1]
\Require 
Source indicator $z \in \{\text{iw}, \text{icl}\}$;
source corpora $\mathcal{D}_{\text{iw}}=\dtrain$, $\mathcal{D}_{\text{icl}}=\dforget$;
WMDP-Cyber questions with gold answers $\mathcal{Q}=\{(X_i,A_i)\}$;
number of retrieved candidates $k_{\text{cand}}$;
number of selected segments $k_z$

\Ensure Segments $\{S_i^z\}$ for each question $X_i$

\State \textbf{Offline indexing:}
\ForAll{document $d \in \mathcal{D}_z$}
    \State Split $d$ into 500-token passages with 50-token overlap $\rightarrow \{p_1, ,\ldots,p_m\}$
    \State Embed each $p_j,j \in [1,m]$ using Qwen Embedding and add to vector index $\mathcal{V}_z$
\EndFor

\State
\State \textbf{Per-question retrieval:}
\ForAll{$(X_i,A_i) \in \mathcal{Q}$}
    \State $q_i \gets \texttt{concat}(X_i,A_i)$ 
    \Comment{Search query}
    \State $P_i^z \gets \texttt{Retrieve}(\mathcal{V}_z,q_i,k_{\text{cand}})$\Comment{Top-$k_{\text{cand}}$ candidates}
    \ForAll{candidate passage $p \in P_i^z$}
        \State $r_i^z(p) \gets \texttt{GPT\text{-}4o}(X_i,A_i,p)$
        \Comment{Score $\in [0,10]$}
    \EndFor
    \State $S_i^z \gets \texttt{Top\text{-}}k_z(P_i^z; r_i^z)$
    \Comment{Select top-$k_z$ passages}
\EndFor
\end{algorithmic}
\end{algorithm}

\paragraph{Step 2: In-context segment retrieval.}
For \ac{icl} segments, 
we retrieve candidate passages from $\dforget$ using the same embedding-based retrieval pipeline.
Since $\dforget$ is never seen during finetuning, 
these passages provide in-context evidence rather than \ac{iw} knowledge.
For each question, 
we retrieve the top-$k_{\text{cand}}$ candidate passages and re-rank them with GPT-4o using the same 10-point scoring prompt.
To keep the number of \ac{icl} and \ac{iw} segments balanced, 
we set $k_{\text{icl}}=k_{\text{iw}}=3$.
We retain the top-$k_{\text{icl}}$ passages as $S_i^{\text{icl}}$ and label them as \ac{icl}.
This is the \ac{icl} instantiation of Algorithm~\ref{algo:segment_retrieval}, 
with $k_{\text{cand}}=10$ and $k_{\text{icl}}=3$.

% \begin{algorithm}[t]
% \caption{ICL segment synthesis}\label{algo:icl_synthesis}
% \begin{algorithmic}[1]
% \Require $\dforget$: cyber-forget corpus (indexed with Qwen Embedding),
%          $\mathcal{Q}$: WMDP-Cyber questions with ground-truth answers,
%          $k_{\text{cand}} = 10$, $K_{\text{icl}} = 3$
% \Ensure ICL segments $\{S^{\text{icl}}_i\}$ for each question $X_i$

% \ForAll{question, choices $(X_i, Y_i) \in \mathcal{Q}$}
%     \State $q_i \gets \texttt{concat}(X_i, Y_i)$
%     \State $\mathcal{C}_i \gets \texttt{Retrieve}(q_i, K_{\text{cand}})$ 
%     \Comment{Top-10 candidates}
%     \ForAll{$p_j \in \mathcal{C}_i$}
%         \State $f_j \gets \texttt{GPT\text{-}4o}(X_i, Y_i, p_j)$ 
%         \Comment{Support score $\in [0, 10]$}
%     \EndFor
%     \State $S^{\text{icl}}_i \gets \texttt{Top\text{-}}K_{\text{icl}}(\mathcal{C}_i, \{f_j\})$ 
%     \Comment{Select top-3 by support score}
% \EndFor
% \end{algorithmic}
% \end{algorithm}

\paragraph{Step 3: Context construction.}
The selected segments $S_i^{\text{iw}}$ and $S_i^{\text{icl}}$ are combined and shuffled to form the final context $C_i = [c_{i,1}, \ldots, c_{i,k}]$.
We use GPT-4o to smooth the concatenated passages into a coherent reference document.
The smoothing prompt enforces five constraints:
\emph{(i)} preserve all factual content,
\emph{(ii)} add transitional phrases,
\emph{(iii)} no new information added,
and \emph{(iv)} maintain original technical terminology.
The resulting context is a single coherent passage; we ensure that the ground-truth provenance labels for each segment are preserved.

\paragraph{Dataset statistics.}
The final WMDP-Cyber++ dataset contains 1{,}987 samples,
with an equal number \ac{iw} and \ac{icl} segments. 
Random source classification achieves AUC of $0.5$.

\section{Evaluation}
\label{sec:experiment}
% Add to preamble: \usepackage[table]{xcolor}
\definecolor{bcsblue}{HTML}{DCEBFA}
\definecolor{hmgreen}{HTML}{DCF0DD}

\begin{table*}[ht]
\small
\centering
\setlength{\tabcolsep}{7pt}
\begin{tabularx}{\textwidth}{l c cccc cccc cccc}
\toprule
&
& \multicolumn{4}{c}{\ac{bcs} ($k{=}1$)}
& \multicolumn{4}{c}{\ac{bcs} ($k{=}2$)}
& \multicolumn{4}{c}{\ac{bcs} ($k{=}3$)} \\
\cmidrule(lr){3-6} \cmidrule(lr){7-10} \cmidrule(lr){11-14}

\textbf{Method} 
& \ac{cac}
& \util  & \rmi  & \ac{bcs}$\uparrow$ & \aps$\uparrow$
& \util & \rmi & \ac{bcs}$\uparrow$ & \aps$\uparrow$
& \util & \rmi & \ac{bcs}$\uparrow$ & \aps$\uparrow$\\
\midrule
\multicolumn{14}{l}{\textbf{LLaMA 3 8B}} \\
\midrule
ContextCite
  & 0.42
  & 0.38 & 0.38 & 0.38 & 0.40
  & 0.38 & 0.38 & 0.38 & 0.40
  & 0.63 & \textbf{0.50} & \textbf{0.56} & \textbf{0.48} \\
TokenShapley
  & 0.43
  & 0.46 & 0.06 & 0.16 & 0.23
  & \textbf{0.59} & 0.19 & 0.33 & 0.37
  & 0.30 & 0.06 & 0.13 & 0.20 \\
AttriBot
  & 0.36
  & 0.25 & 0.38 & 0.31 & 0.33
  & 0.50 & \textbf{0.50} & \textbf{0.50} & \textbf{0.42}
  & 0.50 & 0.38 & 0.43 & 0.39 \\
TracLLM
  & 0.36
  & \textbf{0.50} & \textbf{0.75} & \textbf{0.61} & \textbf{0.45}
  & 0.50 & 0.38 & 0.43 & 0.40
  & \textbf{0.75} & 0.38 & 0.53 & 0.43 \\
\midrule
\multicolumn{14}{l}{\textbf{Qwen 3 8B}} \\
\midrule
ContextCite
  & 0.39
  & 0.53 & 0.21 & 0.33 & 0.36
  & 0.63 & 0.30 & 0.44 & 0.41
  & 0.67 & 0.35 & 0.49 & 0.43 \\
TokenShapley
  & 0.42
  & 0.49 & 0.19 & 0.30 & 0.35
  & 0.63 & 0.21 & 0.36 & 0.39
  & 0.72 & 0.40 & 0.53 & \textbf{0.47} \\
AttriBot
  & 0.34
  & 0.51 & 0.33 & 0.41 & 0.37
  & 0.49 & 0.33 & 0.40 & 0.37
  & 0.53 & 0.33 & 0.42 & 0.38 \\
TracLLM
  & 0.34
  & \textbf{0.86} & \textbf{0.44} & \textbf{0.62} & \textbf{0.44}
  & \textbf{0.79} & \textbf{0.63} & \textbf{0.70} & \textbf{0.46}
  & \textbf{0.86} & \textbf{0.49} & \textbf{0.65} & 0.45 \\
\midrule
\multicolumn{14}{l}{\textbf{Qwen 3 32B}} \\
\midrule
ContextCite
  & 0.56
  & 0.57 & 0.07 & 0.20 & 0.29
  & 0.68 & 0.13 & 0.29 & 0.38
  & 0.78 & 0.19 & 0.38 & 0.45 \\
TokenShapley
  & 0.46
  & 0.62 & 0.05 & 0.17 & 0.25
  & 0.76 & 0.14 & 0.33 & 0.38
  & 0.84 & 0.24 & 0.45 & 0.45 \\
AttriBot
  & 0.47
  & 0.56 & 0.07 & 0.20 & 0.28
  & 0.76 & 0.07 & 0.23 & 0.31
  & 0.78 & 0.24 & 0.43 & 0.45 \\
TracLLM
  & 0.37
  & \textbf{0.86} & \textbf{0.25} & \textbf{0.46} & \textbf{0.41}
  & \textbf{0.94} & \textbf{0.32} & \textbf{0.55} & \textbf{0.44}
  & \textbf{1.00} & \textbf{0.44} & \textbf{0.66} & \textbf{0.47} \\
\bottomrule
\end{tabularx}
\caption{
Attribution methods evaluated using our metrics. 
\ac{cac} is computed with RBO persistence $\rho=0.5$. 
Higher is better. 
\textbf{Bold} denote the best.
Methods cannot disentangle \ac{iw} from \ac{icl} (low \ac{aps}); segment ranking shifts after finetuning (low \ac{cac}).
}
\label{tab:benchmark_cac_bcs}
\end{table*}

\begin{table*}[ht]
\small
\centering
\setlength{\tabcolsep}{5pt}
\begin{tabularx}{\textwidth}{l *{12}{>{\centering\arraybackslash}X}}
\toprule
& \multicolumn{4}{c}{\textbf{TyDiQA}} & \multicolumn{4}{c}{\textbf{HotpotQA}} & \multicolumn{4}{c}{\textbf{CNN/DM}} \\
\cmidrule(lr){2-5} \cmidrule(lr){6-9} \cmidrule(lr){10-13}
\textbf{Method} & Drop@1  & Drop@3  & Drop@5  & LDS  & Drop@1  & Drop@3 & Drop@5  & LDS & Drop@1 & Drop@3 & Drop@5& LDS  \\
\midrule
\multicolumn{13}{l}{\textbf{LLaMA 3 8B}} \\
\midrule
ContextCite
  & \textbf{64.23} & \textbf{113.75} & \textbf{126.62} & \textbf{0.97}
  & \textbf{41.66} & \textbf{76.65} & \textbf{86.78} & \textbf{0.87}
  & {52.79} & \textbf{118.14} & \textbf{155.77} & \textbf{0.94}\\
TokenShapley
  & 11.24 & 35.21 & 47.29 & 0.39
  & 0.88 & 2.45 & 4.13 & 0.12
  & 6.80 & 21.64 & 37.63 & 0.26 \\
Attribot
  & {38.94} & {48.50} & 51.57 & {0.88}
  & {23.29} & {30.25} & 31.75 & {0.76}
  & \textbf{52.48} & {109.36} & {141.15} & {0.85} \\
TracLLM
  & 17.26 & 40.08 & {51.80} & 0.64
  & 22.32 & 30.23 & {32.56} & 0.71
  & 42.76 & 98.09 & 136.30 & 0.82\\
\midrule
\multicolumn{13}{l}{\textbf{Qwen 3 8B}} \\
\midrule
ContextCite
  & {61.66} & 107.86 & 120.02 & \textbf{0.96}
  & 45.20 & 81.74 & 92.24 & \textbf{0.87}
  & 51.86 & 114.49 & 150.49 & \textbf{0.94}
  \\
TokenShapley
  & 49.81 & {113.93} & 140.27 & 0.56
  & 3.47 & 9.88 & 16.72 & 0.12
  & 12.98 & 42.09 & 64.17 & 0.31 \\
Attribot
  & \textbf{99.75} & \textbf{138.81} & {146.37} & {0.94}
  & \textbf{92.24} & \textbf{121.87} & \textbf{126.38} & {0.85}
  & \textbf{104.86} & \textbf{214.41} & \textbf{278.33} & {0.87} \\
TracLLM
  & 50.23 & 119.22 & \textbf{147.34} & 0.76
  & {88.65} & {119.75} & {125.51} & 0.81
  & {87.21} & {190.28} & {256.14} & 0.74\\

\midrule
\multicolumn{13}{l}{\textbf{Qwen 3 32B}} \\
\midrule
ContextCite
  &{40.89} & \textbf{54.89} & \textbf{55.79} & \textbf{0.90}
  & {26.51} & \textbf{37.58} & \textbf{40.06} & \textbf{0.77}
  & {55.90} & {121.75} & {163.64} & \textbf{0.85}\\
TokenShapley
  &6.46 & 30.74 & 46.17 & 0.46
  &1.24 & 3.32 & 5.45 & 0.11 
  &14.48 & 42.19 & 69.56 & 0.33\\
Attribot
 &\textbf{41.05} & {52.97} & 54.60 & {0.88}
  & \textbf{27.02} & {36.25} & 38.16 & {0.75}
  & \textbf{64.02} & \textbf{130.94} & \textbf{168.55} & {0.84}\\
TracLLM
  &19.62 & 45.09 & {55.16} & 0.67
  & 25.79 & 35.80 & {38.50} & 0.73
  & 49.12 & 111.31 & 151.46 & 0.71 \\
\bottomrule
\end{tabularx}
\caption{
Attribution performance measured using Drop@$k$ and LDS. 
Higher is better. \textbf{Bold} denotes the best result.
Both metrics are mostly stable and agree between the settings.
No method is consistently the best. 
However, Drop@$k$ and LDS tend to favor LOO-based methods, 
highlighting the need for complementary metrics such as \ac{aps}.
}
\label{tab:benchmark_attribution}
\end{table*}

\subsection{Experimental setup}
\label{subsec:experimental_setup}
\paragraph{Implementation details.}
All experiments are conducted on two NVIDIA H100 GPUs.
We evaluate attribution methods using three open weight \acp{llm}: 
LLaMA3-8B~\cite{grattafiori2024llama3herd}, 
Qwen3-8B~\cite{yang2025qwen3technicalreport}, 
and Qwen3-32B~\cite{yang2025qwen3technicalreport}.
We use greedy decoding throughout to ensure reproducible results.

\paragraph{Attribution methods.}
We evaluate four representative methods:
ContextCite~\cite{cohen-wang2024contextciteattributingmodel} fits a sparse linear surrogate model over perturbed context masks to estimate segment contributions. 
TokenShapley~\cite{xiao2025tokenshapleytokenlevel} achieves fine-grained, 
token-level attribution by combining Shapley-style values with KNN-based retrieval. 
AttriBoT~\cite{liu2024attribotbagtricks} focuses on efficiency by directly approximating \ac{loo} context attribution.
TracLLM~\cite{wangtracllmgenericframework} scales perturbation-based traceback to long contexts by combining informed search with \ac{loo}-style scores.

\paragraph{Metrics.}
We rely on the evaluation metrics introduced in this work:
\ac{bcs},
\ac{cac},
\ac{aps},
and \ac{ssp}.
For \ac{cac}, 
we compute ranking consistency using \ac{rbo} with persistence $\rho=0.5$,
which is a strongly top-heavy setting.
Thus, \ac{cac} primarily measures whether the base and finetuned models agree on the most highly attributed context segments.
For comparison with prior likelihood-based evaluations, 
we also report \emph{top-$k$ log-probability drop} and \ac{lds}, following ContextCite.

\paragraph{Adapting methods for source separation.}
Since existing attribution methods output a scalar contributive score per context segment, 
we adapt them with a disentanglement interface $\interface$ that uses the contributive score to predict \ac{iw}/\ac{icl} source labels,
% which is applied on the finetuned model only. 

For ContextCite and TracLLM,  we contrast each segment's normalized positive score against a no-context baseline, 
which captures how much of the response is recoverable from \ac{iw} alone,
segments whose contributive score dominates this baseline are labeled \ac{iw} and \ac{icl}. 
For AttriBoT and TokenShapley, 
we normalize the scores into mass fractions and label a segment \ac{icl} when it carries an above uniform share of the attribution mass, and \ac{iw} otherwise. 

Both rules follow the same intuition: a relevant segment whose removal barely affects the response is presumed covered by \ac{iw} knowledge. 
Each adaptation requires at most one additional forward pass. 
Further details are provided followed by each method in the appendix.

% TODO
% Further details are provided in appendix X/ Supplementary material.

\paragraph{Datasets.}
To measure \ac{iw}/\ac{icl} disentanglement, 
we use our \wmdppp dataset.
To evaluate attributive contribution we use:
the TyDiQA~\cite{clark2020tydiqabenchmark} validation set (5,077),
the HotpotQA~\cite{yang2018hotpotqadatasetdiverse} validation set (7,410), 
and the validation set (13,368) of CNN/Daily Mail dataset~\cite{see2017getpointsummarization}.
We also use validation subset (1,000) of MS-marco dataset~\cite{bajaj2016msmarcohuman},
and subset of training set (1,000) of NQ dataset~\cite{kwiatkowski2019naturalquestionsbenchmark},
which are reported as additional results in the appendix.

% The top-$k$ log-probability drop measures how much the log-probability of the original response decreases after removing the top-$k$ attributed context segments, 
% where a larger drop indicates stronger contextual influence.
% LDS measures how well the attribution scores predict the actual effect of ablating context pieces, with a higher score indicating better attribution fidelity.

\subsection{Experimental results}

\paragraph{Controlled knowledge exposure affects attribution.}
Table~\ref{tab:benchmark_cac_bcs} reports the results on \wmdppp,
including \ac{bcs} across $k \in \{1,2,3\}$, 
\ac{cac} at \ac{rbo} persistence $\rho=0.5$, 
and their harmonic mean \ac{aps}. 

Across all settings,
\ac{cac} remains low ($0.34$--$0.56$), 
showing that attribution rankings change after controlled
knowledge exposure. 
% \ac{bcs} and \ac{cac} can also diverge substantially.
TracLLM attains the highest \ac{bcs} in seven out of nine model--$k$ settings
but has among the lowest \ac{cac}. 
In contrast, TokenShapley achieves higher \ac{cac} on the 8B models, 
yet low \ac{bcs} in some settings (e.g., $0.16$ at $k{=}1$ on LLaMA3-8B). 
\ac{aps} makes this trade-off explicit: 
on Qwen3-8B at $k{=}3$, 
TokenShapley exceeds TracLLM in \ac{aps} ($0.47$ vs.\ $0.45$) 
despite lower \ac{bcs} ($0.53$ vs.\ $0.65$), 
owing to its higher \ac{cac}.
Nevertheless, averaged across settings, 
TracLLM performs best under our task-level \ac{bcs} evaluation and achieves the strongest average \ac{aps}, 
although its lower \ac{cac} indicates less stable rankings after controlled knowledge exposure.
We conclude that attribution rankings depend both on the provided context
and the model's knowledge.

\begin{figure}[t]
    \centering
    \includegraphics[width=0.47\textwidth]{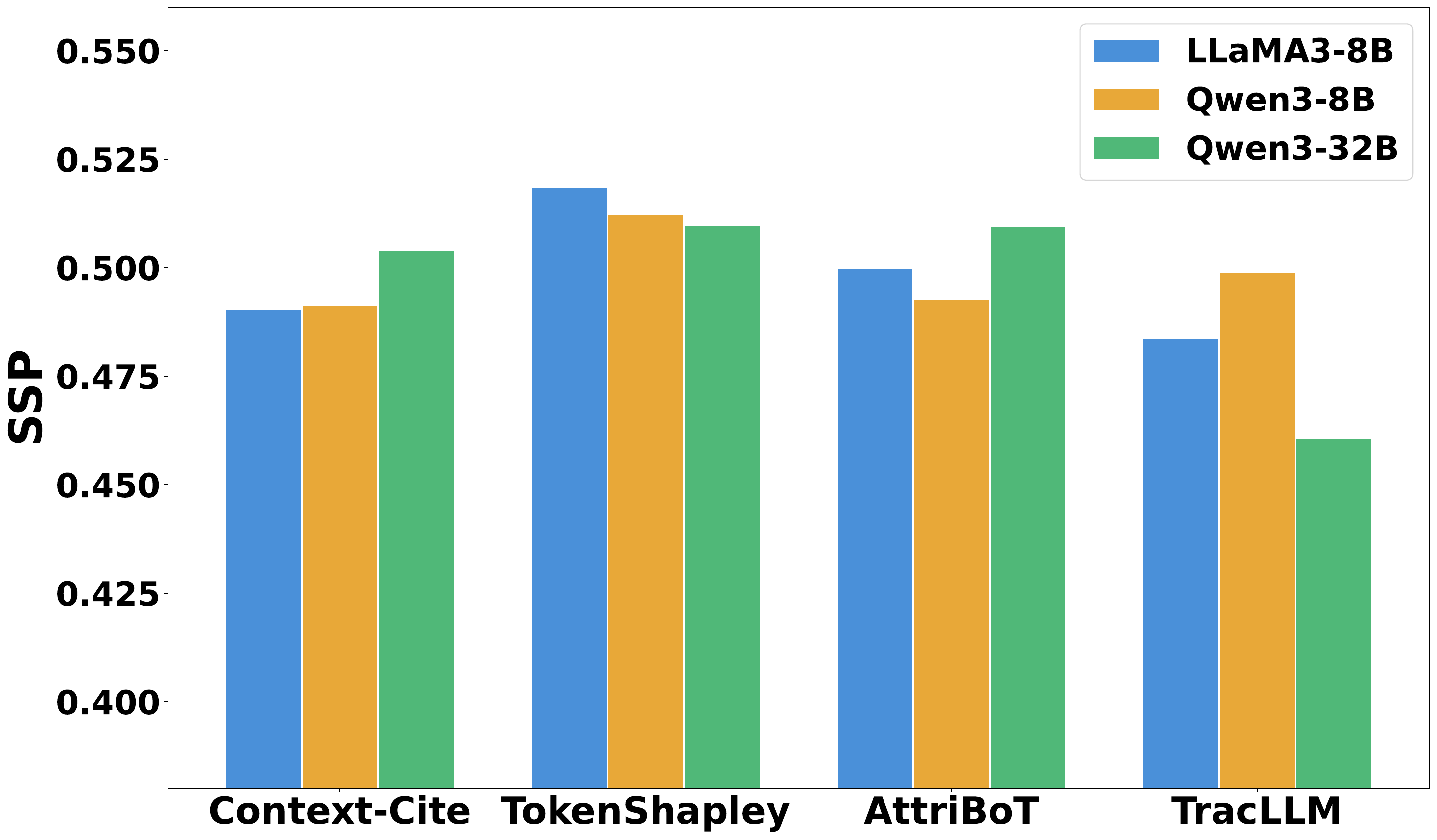}
    \caption{\ac{ssp} on \wmdppp dataset across different attribution methods on LLaMA3-8B, Qwen3-8B, and Qwen3-32B. All of the attribution methods with the disentanglement adaptation achieve near-random results at about 0.5.}
    \label{fig:ssp_demonstration}
\end{figure}

\paragraph{Attribution does not reflect disentanglement.}
Figure~\ref{fig:ssp_demonstration} shows \ac{ssp} on \wmdppp, 
where each segment's provenance (\ac{iw} vs.\ \ac{icl}) is known by construction. 
In all evaluated settings, 
\ac{ssp} stays close to chance.
The best method, TokenShapley, reaches only $0.52$ for LLaMA3-8B and lower for Qwen models.
No method approaches a usable level.
Hence, attribution scores do not reflect whether the segments contribute based on \ac{icl} or \ac{iw}.

%  ########### old version source with too many digits ##########
% \paragraph{Source Separation Results.} 
% Figure~\ref{fig:ssp_demonstration} illustrates the source separation precision of context attribution methods across three models,
% LLaMA3-8B~\cite{grattafiori2024llama3herd}, 
% Qwen3-8B~\cite{yang2025qwen3technicalreport}, 
% and Qwen3-32B~\cite{yang2025qwen3technicalreport}. 
% Among all methods, 
% TokenShapley achieves the highest separation, 
% with $\sspicl =0.636$ and $\sspiw=0.510$ on LLaMA3-8B, 
% ($0.589$, $0.507$) on Qwen3-8B, 
% and ($0.558$, $0.505$) on Qwen3-32B. 
% Notably, 
% the consistent gap between $\sspicl$ and $\sspiw$ within TokenShapley suggests that with the previously proposed approaches, 
% in-context contributions are slightly easier to identify than \ac{iw}, 
% yet neither reaches a practically reliable level. 
% This suggests that more principled surrogate formulations or the additional classifier are necessary to address source disentanglement and handle the context, 
% which is from the \ac{iw} data.

% \begin{figure}
%     \centering
%     \includegraphics[width=1\linewidth]{figures/rank_heatmap.pdf}
%     \caption{ranking heatmap}
%     \label{fig:rank_heatmap}
% \end{figure}

\paragraph{Likelihood-based metrics are insufficient.}
Table~\ref{tab:benchmark_attribution} reports Drop@$k$ and \ac{lds}.
The two metrics produce similar rankings across datasets, 
suggesting that they capture a shared likelihood-sensitivity signal.
This is expected because several evaluated methods are closely aligned with these metrics by construction.
For example, Drop@1 is closely aligned with \ac{loo}-based attribution, since both measure the effect of removing individual context segments on the model's likelihood.
Similarly, \ac{lds} is closely aligned with ContextCite-style perturbation objectives, 
as both evaluate whether attribution scores explain likelihood changes under context perturbations.
As a result, 
likelihood-based evaluations tend to favor methods whose scoring mechanisms match their own perturbation-based assumptions.

In contrast, 
\ac{aps} yields different rankings, 
especially on the two Qwen models,
indicating that task-level attribution quality is not fully captured by likelihood-based metrics.
This does not make \ac{aps} a universal metric.
Instead, \ac{aps} serves as a complementary metric by jointly capturing task-level attribution quality and consistency under controlled knowledge exposure.
\ac{aps} yields substantially different method rankings from Drop@$k$ and \ac{lds},
including different best-performing methods on the two Qwen models.
These ranking differences are further visualized in the appendix.
Rather, the divergence highlights the need for complementary evaluation:
likelihood-based metrics alone cannot assess attribution behavior under \ac{iw} overlap.

\begin{figure*}[!t]
    \centering
    \includegraphics[width=0.9\textwidth]{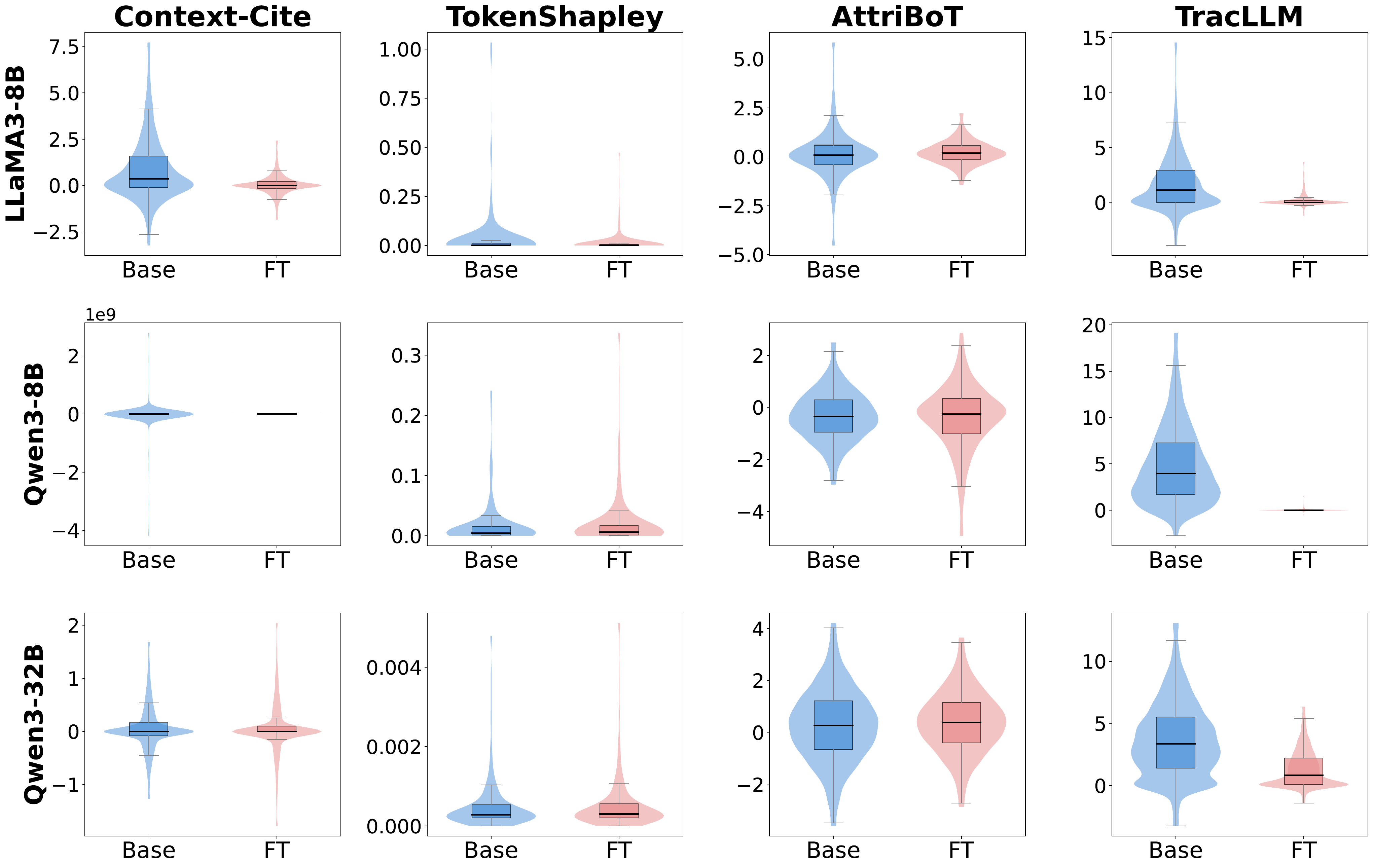}
    \caption{Contributive score distribution shift of different attribution methods on two LLaMA3-8B, Qwen3-8B, and Qwen3-32B. TracLLM and ContextCite show the most significant contributive score distribution shift.}
    \label{fig:distribution_shift}
\end{figure*}

\paragraph{Score changes based on the \attr mechanism.} 
Figure~\ref{fig:distribution_shift} illustrates that the distribution of the attribution scores shifts from the base model to the fine-tuned model.
This pattern reveals that the attribution methods assign different attribution
scores based on the \attr mechanism.
With ContextCite, AttriBoT, and TracLLM, which assign the score to each segment through the response's log-probability, 
the attribution scores shift toward lower values after fine-tuning.
On the other hand, 
TokenShapley computes attribution scores for each segment from the hidden-state representations of the \ac{llm} rather than the output likelihood, 
and its distribution changes only slightly.
These results suggest that neither attribution mechanism produces attribution scores that remain consistent and robust across knowledge-source conditions.

% \paragraph{Model size leads to score distribution shift.}
% Figure~\ref{fig:distribution_shift} shows that finetuning induces distributional changes in the attribution scores for the 8B models; 
% this effect is reduced for Qwen3-32B. 
% For LLaMA3-8B and Qwen3-8B, 
% several methods, 
% especially TracLLM, 
% show a clear compression of scores after fine-tuning.
% Distributions are narrower and closer to zero. 
% This pattern suggests that 
% after the model acquires the knowledge, 
% it requires less support from the provided context. 

% On the other hand,  the Qwen3-32B results show minor changes for
% ContextCite,  AttriBoT, and TokenShapley. 
% The base and finetuned
% score distributions remain largely overlapping.  
% TracLLM still exhibits a noticeable shift on, 
% although it is less pronounced than for the 8B models. 

% We conclude that neither simple linear surrogate models nor heuristic confidence-based scoring functions are expressive enough to produce attribution scores that are consistent across knowledge-source conditions.
% These results motivate the need for an attribution framework that explicitly disentangles in-context from \ac{iw} contributions.

% indicating that the larger model's attribution behavior is less affected by finetuning. 

\section{Related work}
\label{sec:related_work}

It was shown that \acp{llm} generate responses based on the information from the training data (\ac{iw}), provided in the context (\ac{icl}), 
or the mixture of the two~\cite{chan2022transformersgeneralizedifferently,tao_when_2024,zhao_understanding_2025}. 
Identifying which of these sources drives the response is crucial to interpreting hallucinations~\cite{du2024contextpriorknowledge, tao2024whencontextleads, chuang2024lookbacklensdetecting,kim_how_2026}. 
ContextCite~\cite{cohen-wang2024contextciteattributingmodel} hints that ablation-based methods can assign low scores to relevant context when the model relies on \ac{iw} knowledge.
However, a systematic analysis of this phenomenon is not acknowledged in other work~\cite{horovicz2024tokenshapinterpretinglarge,
xiao2025tokenshapleytokenlevel,
liu2024attribotbagtricks,
wangtracllmgenericframework}. 
To address this gap, AttriWiki~\cite{brink2026probingknowledgeattribution} uses a lightweight classifier to estimate contributive scores. 
Despite promising initial results for context attribution, 
they fail when handling the context that is in the training data~\cite{cohen-wang2024contextciteattributingmodel}. 
Furthermore, 
they cannot disentangle data-source usage within the input context, 
leading to ambiguous attribution outcomes. 
To study this challenge, 
we introduce an evaluation protocol to assess the accuracy of context attribution when disentangling in-weight and in-context contributions.

\section{Discussion}
\label{sec:discussion}

\paragraph{Weight access.}
Our evaluation requires white-box access to the model.
It relies on comparing measurements from the base and finetuned models.
This is justified because our goal is to evaluate the attribution methods,
not to compete with them in a post-hoc setting.
In a deployment setting, the end-user will use an attribution method that was,
by design, evaluated using our protocol.

% \paragraph{Score suppression is uncorrelated with memorization}
% Figure~\ref{fig:score_supression} demonstrates that attribution distortion scales with how strongly the finetuned model has memorized the relevant knowledge. 
% These tests are conducted by regressing per-segment score suppression against memorization
% confidence for all \ac{iw} segments. 
% From the results, 
% three of the four methods exhibit near-zero coefficients of determination, 
% indicating that memorization strength has virtually no predictive power over attribution distortion. 
% TracLLM is an exception with a weak but statistically positive correlation, 
% suggesting partial sensitivity to parametric recall in its \ac{loo} formulation. 
% \emph{
% \textbf{These results imply that the distortion introduced by \acs{iw} overlap is not a smooth, 
% predictable degradation amenable to post-hoc correction, 
% but rather an unpredictable perturbation that requires explicit disentanglement.}
% }

\paragraph{Contributive vs. corroborative attribution.}
Prior work has defined contributive and corroborative attribution
\cite{worledge2023unifyingcorroborativecontributive}.
Contributive attribution quantifies how important a source, 
such as a training data sample, is to \model,
and is usually measured by its counterfactual contribution to the loss or output.
Corroborative attribution has also been referred to as citation in prior work
~\cite{nakano2022webgptbrowserassistedquestionanswering,menick2022teachinglanguagemodels},
and can be measured by exact match, valid paraphrase, or textual entailment.

ContextCite formalizes context attribution as contributive context attribution
\cite{cohen-wang2024contextciteattributingmodel},
which quantifies the contribution of each context segment to the model's response.
We adopt this definition throughout the paper
and further interpret context attribution at two levels:
the text-level effect of a context segment, 
and the knowledge-level role of the segment.
This distinction is important because \ac{iw} knowledge can mask the observable effect of a context segment,
which helps us interpret the contributive score in a comprehensive way.

\paragraph{Disentanglement is necessary for usable attribution.}
Attribution score is a property of the context–model pair.
Existing methods compute counterfactual effects conditioned on a fixed model;
resulting scores are silently affected by the \ac{iw} knowledge. 
Low score does not guarantee low contribution when the segment knowledge is present in the weights.
This approach is potentially misleading, 
particularly when attribution scores are treated as direct evidence, 
as seen in the citation rewards mechanism utilized in SelfCite
\cite{chuang2025selfciteselfsupervisedalignment}.
Our analysis shows that this limitation is not from one method, but arises broadly from existing context attribution methods under the overlap between \ac{icl} and \ac{iw} knowledge.

\paragraph{Removal primitive.}
Our findings reflect that all evaluated methods rely on text removal as the primitive:
leave-one-out (AttriBot, TracLLM)
random-ablation surrogates (ContextCite),
or Shapley-style marginal contributions (TokenShapley).
The primitive ignores \ac{iw} knowledge by design,
since removing a context segment does not remove the learned information.
Future attribution methods should quantify both the contribution of a segment in the context and from the weights.
Our protocol and \wmdppp aid the design of such methods.

\paragraph{Finetuning as a proxy for in-weight knowledge.}
Our protocol instantiates IW knowledge to the \acp{llm} through LoRA 
finetuning on \dtrain. The reason for this choice due to the uncontrollable and no public training data information of the public \acp{llm}. 
Therefore the fine-tuning allow us to control that the knowledge is actually acquired to \acp{llm}. 
For the further works, we suggest the ablation for the behaviour of attribution methods across different finetuning approaches and different hyperparameters of the LoRA.
\section{Conclusion}
\label{sec:conclusion}
In this work, we show that existing attribution methods fail when the context overlaps with the \ac{iw} knowledge.
We introduce an evaluation protocol relying on new metrics (\ac{bcs}, \ac{cac}, \ac{aps}, \ac{ssp}),
and a benchmark with controlled \ac{iw}/\ac{icl} provenance labels (\wmdppp). 
We demonstrate that
\begin{enumerate*}[label=\arabic*)]
    \item high ranking consistency does not imply faithful attribution;
    \item all methods perform near chance at source separation under a source-labeling interface. 
\end{enumerate*}
Our findings call for attribution methods that can disentangle \ac{icl}/\ac{iw}contributions.

% \paragraph{Limitations and future work.} 
% We diagnose the problem but do not propose a method
% that can attribute contributive scores and identify each segment's knowledge source; 
% developing such a framework is the central open direction motivated by our findings.
% \newpage
\bibliography{references}
\clearpage
\appendix

% \section{Notation}
\section{Additional results}

\subsection{Additional results on MS-Marco and NQ dataset}
we extend the attribution experiments from table~\ref{tab:benchmark_attribution} to two additional datasets: 
MS-MARCO and Natural Questions (NQ). 
Table~\ref{tab:benchmark_attribution_appendix} reports the results. 
The trends for the benchmark are consistent with the main evaluation: 
ContextCite achieves the highest \ac{lds} across both datasets and models, 
confirming its strong linear faithfulness, 
while AttriBot leads on Drop@1 due to its \ac{loo}-based formulation. 
TokenShapley remains the weakest performer overall, 
particularly on LLaMA3-8B where its Drop@1 falls to 3.22 (MS-MARCO) and 12.39 (NQ). 
These additional results reinforce our finding that method rankings are model-dependent. 
AttriBot and ContextCite perform comparably on LLaMA3-8B but diverge on Qwen3-8B, and that high removal impact does not necessarily entail high linear faithfulness.
\begin{table*}[h!]
\small
\centering

\setlength{\tabcolsep}{4pt}
\begin{tabular}{l cccc cccc}
\toprule
& \multicolumn{4}{c}{\textbf{MS-MARCO}} & \multicolumn{4}{c}{\textbf{NQ}} \\
\cmidrule(lr){2-5} \cmidrule(lr){6-9}
\textbf{Method} & Drop@1 $\uparrow$ & Drop@3 $\uparrow$ & Drop@5 $\uparrow$ & LDS $\uparrow$ & Drop@1 $\uparrow$ & Drop@3 $\uparrow$ & Drop@5 $\uparrow$ & LDS $\uparrow$ \\
\midrule
\multicolumn{9}{l}{\textbf{LLaMA 3 8B}} \\
\midrule
ContextCite
  & 47.51 & \textbf{88.02} & \textbf{102.99} & \textbf{0.83}
  & 34.07 & \textbf{49.43} & \textbf{54.02} & \textbf{0.89} \\
TokenShapley
  & 3.22 & 9.97 & 17.23 & 0.16
  & 12.39 & 37.40 & 49.82 & 0.43 \\
Attribot
  & \textbf{49.91} & 84.26 & 96.30 & 0.82
  & \textbf{34.21} & 47.10 & 52.98 & 0.86 \\
TracLLM
  & 46.04 & 84.32 & 100.51 & 0.80
  & 14.31 & 39.77 & \textbf{54.02} & 0.62 \\
\midrule
\multicolumn{9}{l}{\textbf{Qwen 3 8B}} \\
\midrule
ContextCite
  & 64.43 & \textbf{113.14} & \textbf{130.18} & \textbf{0.84}
  & 92.12 & \textbf{135.78} & \textbf{144.48} & \textbf{0.94} \\
TokenShapley
  & 4.80 & 12.52 & 22.12 & 0.14
  & 48.35 & 111.84 & 136.42 & 0.58 \\
Attribot
  & \textbf{67.73} & 108.94 & 123.42 & 0.83
  & \textbf{92.67} & 133.16 & 143.04 & 0.93 \\
TracLLM
  & 60.98 & 106.13 & 123.80 & 0.81
  & 44.16 & 113.24 & 144.29 & 0.75 \\
\midrule
\multicolumn{9}{l}{\textbf{Qwen 3 32B}} \\
\midrule
ContextCite
  & 18.31 & \textbf{34.01} & \textbf{40.13} & \textbf{0.77}
  & 34.24 & \textbf{49.18} & \textbf{51.50} & \textbf{0.91} \\
TokenShapley
  & 1.74 & 4.27 & 7.34 & 0.15
  & 15.70 & 38.99 & 47.76 & 0.52 \\
Attribot
  & \textbf{19.12} & 31.19 & 36.27 & 0.72
  & \textbf{34.42} & 47.90 & 50.64 & 0.89 \\
TracLLM
  & 16.11 & 30.09 & 36.69 & 0.68
  & 14.54 & 39.04 & 51.31 & 0.68 \\
\bottomrule
\end{tabular}
\caption{
Attribution performance measured using Drop@$k$ and LDS. 
Higher is better. \textbf{Bold} denotes the best result.
Both metrics are mostly stable and agree between the settings.
No method is consistently the best. 
However, Drop@$k$ and LDS tend to favor LOO-based methods, 
highlighting the need for complementary metrics such as \ac{aps}.
}
\label{tab:benchmark_attribution_appendix}
\end{table*}

\subsection{ROC curves for the disentanglement of the \ac{iw} and \ac{icl} sources}

\begin{figure}[!ht]
    \centering
    \includegraphics[width=0.47\textwidth]{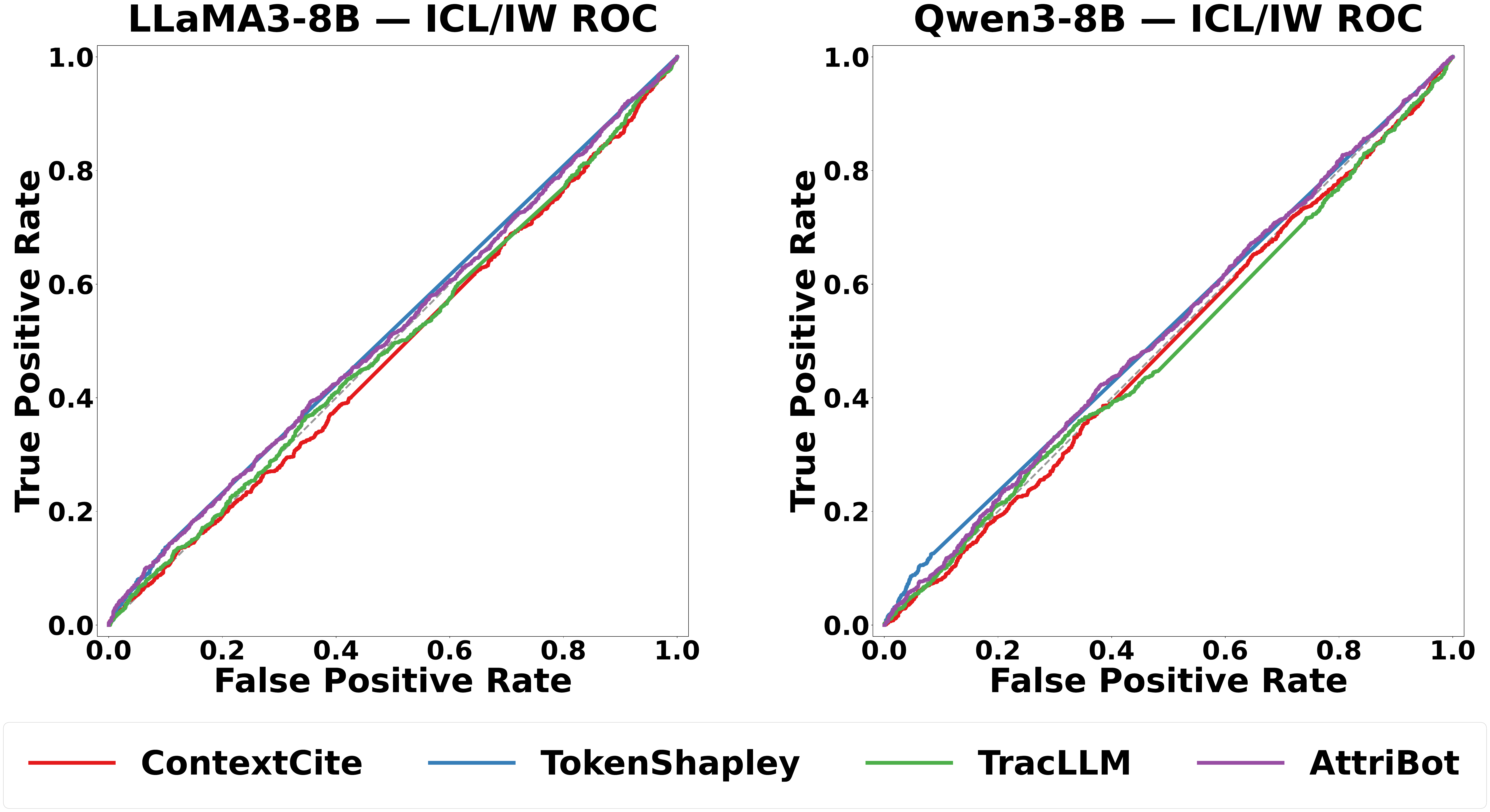}
    \caption{ROC curves for classifying context segments as ICL
(top) or IW (bottom) using attribution scores on LLaMA3-8B
(left) and Qwen3-8B (right). All methods achieve AUC $\approx$
0.5, confirming that attribution scores carry no discriminative
signal for knowledge provenance at any threshold.}
    \label{fig:ssp_roc}
    \vspace{-3mm}
\end{figure}

Figure~\ref{fig:ssp_roc} presents ROC curves for classifying context segments as \ac{icl} or \ac{iw} using each method's attribution scores. 
All methods yield AUC values nearly the random baseline (0.5) on both models and both detection tasks, with curves closely tracking the diagonal. 
This result suggests that recent attribution methods, 
with the interface, have challenge in disentangling the \ac{icl} and \ac{iw}. 
Therefore, a better attribution method should be proposed to both return the contribution score and do the disentanglement.

\subsection{CAC and BCS Results under different RBO persistence parameters}
\label{app:cac-rbo-pvariation}
\begin{figure}[h!]
  \centering
  \includegraphics[width=0.47\textwidth]{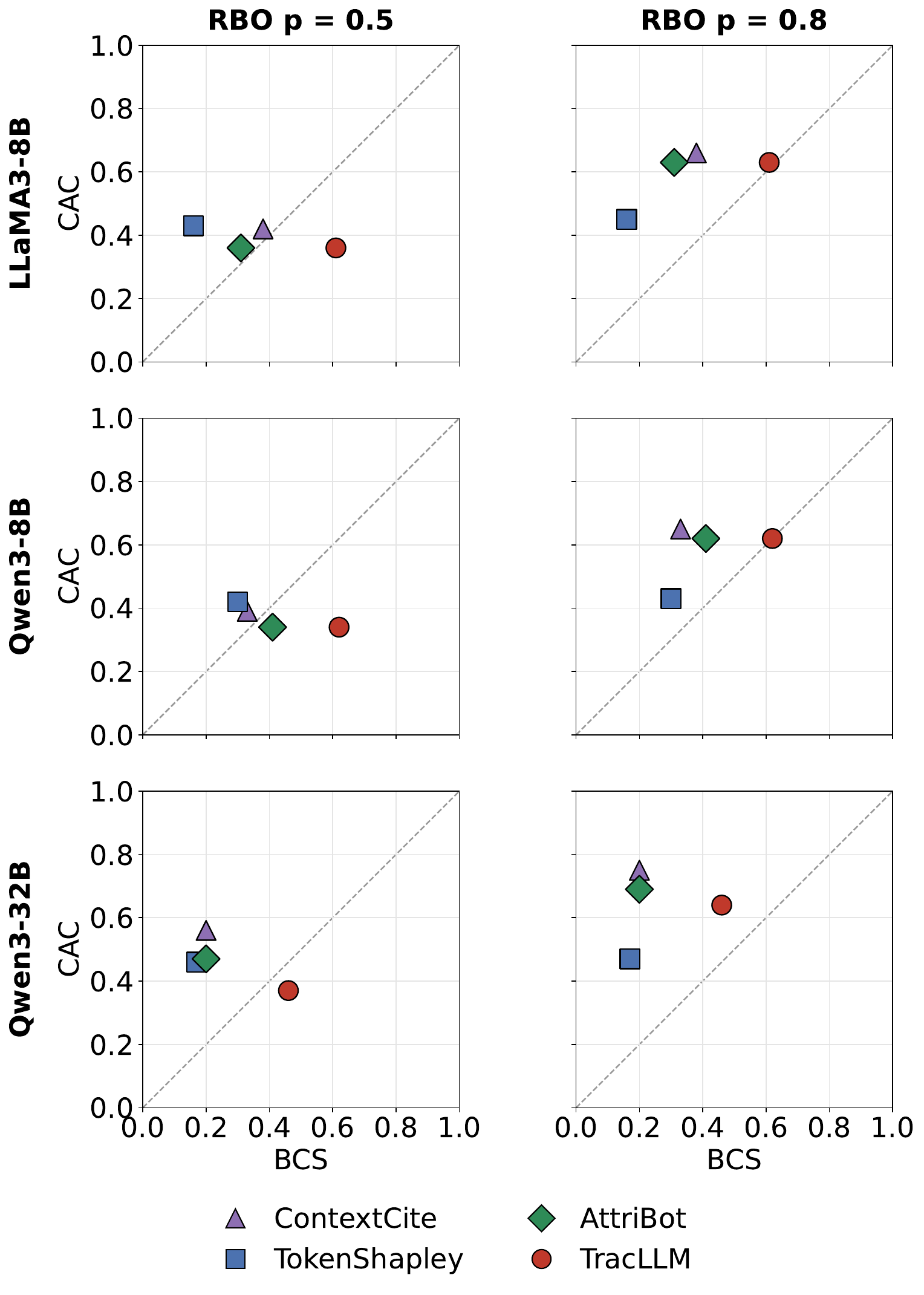}
  \caption{
  \textbf{CAC vs.\ BCS} for four attribution methods across three target models.
  BCS measures task-level attribution quality at $k{=}1$, while
  CAC measures ranking consistency between the base and finetuned models
  using RBO with $\rho{=}0.5$ and $\rho{=}0.8$.
  The dashed line denotes $y{=}x$.
  }
  \label{fig:cac_bcs_pvariation}
\end{figure}

This part provides supplementary results on CAC and BCS
under different choices of the RBO persistence parameter $\rho$.
In the main text, CAC is computed with $\rho{=}0.5$,
which places greater emphasis on agreement among the highest-ranked
context segments.
Since $\rho$ controls how strongly RBO down-weights deeper ranks,
we additionally report results for $\rho{=}0.8$,
corresponding to expected evaluation depths of approximately $2$ and $5$,
respectively.
Figure~\ref{fig:cac_bcs_pvariation} shows the resulting CAC--BCS patterns
across all three target models.

The relative comparison between attribution methods remains largely stable across the two values of $\rho$.
Although the absolute CAC values change as deeper ranks receive greater weight,
the overall CAC--BCS patterns are preserved.
Notably, methods with stronger BCS do not necessarily achieve higher CAC,
and vice versa, showing that neither metric alone provides a complete comparison of attribution methods.
This observation motivates \ac{aps}, 
which provides a joint summary of performance across both criteria.
The persistence of these patterns across different values of $\rho$ further suggests that our main findings do not depend critically on the specific choice of $\rho=0.5$.

\subsection{Ranking differences across evaluation metrics}
\label{app:metric_ranking_comparison}
\begin{figure}[t]
    \centering
    \includegraphics[width=\linewidth]{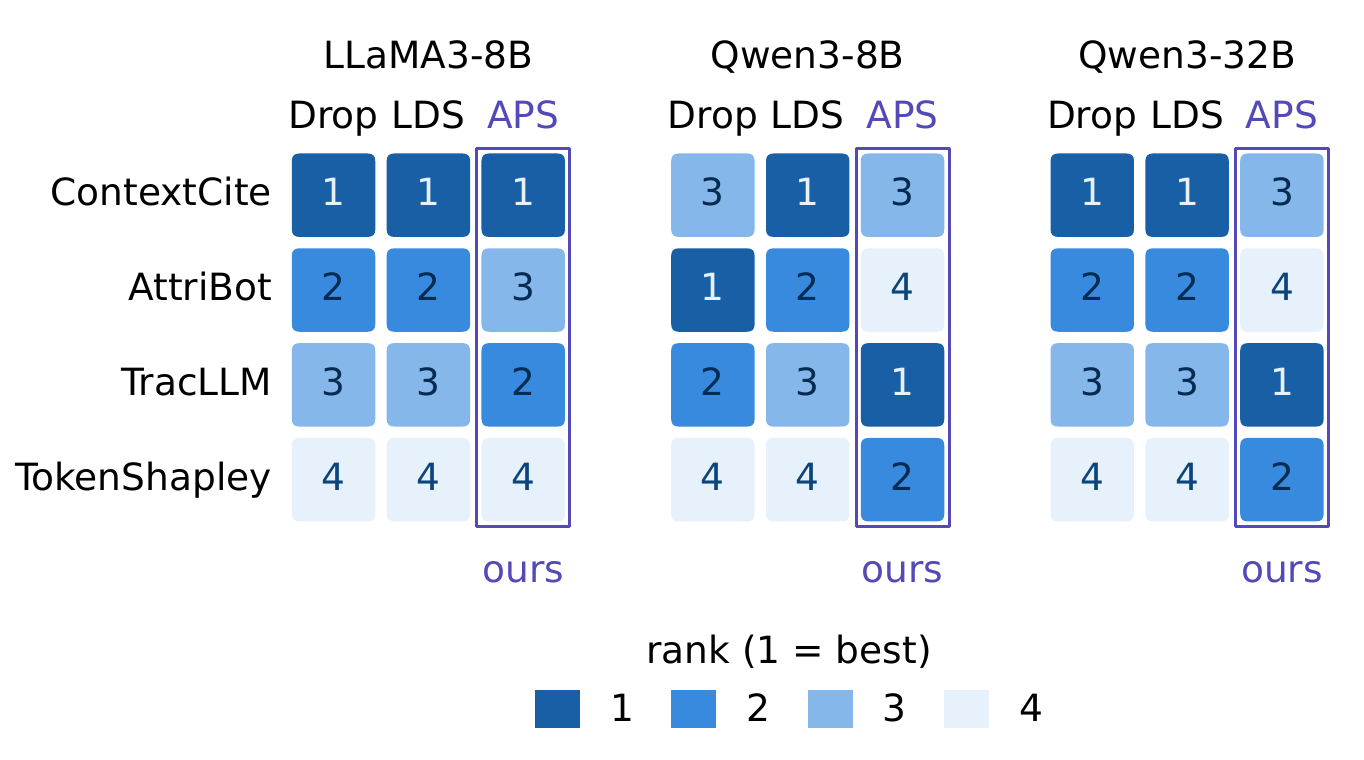}
    \caption{
    Ranking of attribution methods under Drop@$k$, \ac{lds}, and \ac{aps} across the evaluated models.
    Numbers denote ranks, with 1 indicating the best-performing method.
    \ac{aps} additionally captures task-level attribution quality and consistency under controlled knowledge exposure.
    }
    \label{fig:metric_ranking_comparison}
\end{figure}

Figure~\ref{fig:metric_ranking_comparison} compares attribution-method rankings under
Drop@$k$, \ac{lds}, and \ac{aps}.
Drop@$k$ and \ac{lds} produce largely similar rankings, whereas \ac{aps} changes the relative ordering of methods, particularly on the two Qwen models.
This further shows that these metrics capture complementary aspects of attribution quality.

\section{Additional implementation details for the fine-tuning large language model}
For \ac{llm} fine-tuning, 
we fine-tune the models on the WMDP dataset, using the cyber retain subset. 
We employ LoRA with rank $r=16$ and scaling factor $\alpha=64$. 
The adapted modules differ by architecture: 
for LLaMA3-8B, 
we target the query, key, value, and output projections, 
whereas for Qwen3 models, 
we additionally include the gate and down projections. 
Both models are trained with a batch size of 4, 
gradient accumulation is 8, 
and using the AdamW optimizer with a learning rate of $0.0001$. 
The best checkpoints are selected after three epochs of full supervised finetuning. 

Furthermore, 
to verify that the training data is non-trivial with respect to the base model's pre-training corpus—and that the finetuned model has genuinely acquired knowledge from the training set, 
we evaluate both the base and finetuned models in a no-context setting with the multiple-choice accuracy metric. 
Without any retrieved context, 
the base models achieve $42.2\%$ (LLaMA3-8B) and $53.8\%$ (Qwen3-8B), 
whereas the fine-tuned model reach $76.6\%$ and $96.8\%$, 
respectively. 
The substantial performance gap confirms that the fine-tuned models have effectively learned the target knowledge, 
validating their suitability for our evaluation protocol.

\section{Extending attribution methods for source disentanglement}
\label{subsec:extensions_attr}

Existing context attribution methods produce a single importance score per segment,
without distinguishing whether a segment's influence arises from \ac{icl} or from knowledge already encoded in the model's parameters (\acs{iw}).
We extend four representative methods---
ContextCite, 
TokenShapley,
AttriBoT,
and TracLLM---
with a unified disentanglement interface.
Each extension uses only a single model (the fine-tuned model $p_{\theta_{\text{ft}}}$) at inference time,
requiring no access to the base model $p_{\theta_{\text{base}}}$,
making them deployable in practice.

\subsection{ContextCite}

\paragraph{Original attribution method.}
ContextCite estimates the contribution of each context unit by fitting a sparse
linear surrogate over randomly ablated contexts. 
Given the context $C=\{c_1,\ldots,c_k\}$, 
we sample $M$ binary masks $\mathbf{m}^{(j)}\in\{0,1\}^{k}$, 
where each context unit is retained independently with probability $0.5$. 
For each mask, the method computes the log-probability of the response under the corresponding ablated context. 
The attribution coefficients are then obtained by solving the LASSO objective:
\begin{equation}
\begin{aligned}
\hat{\boldsymbol{\beta}}
=
\arg\min_{\boldsymbol{\beta}}
\Bigg\{
&\sum_{j=1}^{M}
\Bigl(
\log p_{\theta_{\mathrm{ft}}}
\bigl(
Y \mid (C \odot \mathbf{m}^{(j)}) \oplus X
\bigr)
\\
&\qquad
- \boldsymbol{\beta}^{\top}\mathbf{m}^{(j)}
\Bigr)^2
+ \lambda \lVert \boldsymbol{\beta} \rVert_1
\Bigg\}.
\end{aligned}
\label{eq:contextcite}
\end{equation}
% \begin{equation}
% \begin{aligned}
% \hat{\boldsymbol{\beta}}
% =
% \arg\min_{\boldsymbol{\beta}}
% &\sum_{j=1}^{M}
% \left(
% \log p_{\theta_{\mathrm{ft}}}
% \bigl(
% Y \mid (C \odot \mathbf{m}^{(j)}) \oplus X
% \bigr)
% -
% \boldsymbol{\beta}^{\top}\mathbf{m}^{(j)}
% \right)^2 \\
% &\quad + \lambda \lVert \boldsymbol{\beta} \rVert_1,
% \end{aligned}
% \label{eq:contextcite}
% \end{equation}

where $C\odot\mathbf{m}^{(j)}$ denotes the context obtained by retaining the units whose corresponding mask entries are one. 
The attribution score assigned to context unit $c_i$ is
\begin{equation}
    s_i = \hat{\beta}_i \in \mathbb{R}.
\end{equation}

\paragraph{Disentanglement extension.}
We adapt the contributive scores to predict whether each context unit is
associated with \ac{icl} or \ac{iw} knowledge. Let
$s_i^{+}=\max(s_i,0)$ denote the positive contribution of $c_i$, and let
$p_{\emptyset}\in[0,1]$ denote the no-context recovery score of
$p_{\theta_{\mathrm{ft}}}$. This score captures the model's ability to produce
the response using parametric knowledge alone. We define the \ac{icl} proxy
score as
\begin{equation}
s_i^{\mathrm{ICL}}
=
\begin{cases}
\dfrac{s_i^{+}}{\max_{j} s_j^{+}},
& \text{if } \max_{j}s_j^{+}>0, \\[7pt]
0,
& \text{otherwise},
\end{cases}
\label{eq:icl_alloc}
\end{equation}
and compute the corresponding \ac{iw} proxy score as
\begin{equation}
    s_i^{\mathrm{IW}}
    =
    p_{\emptyset}
    \bigl(1-s_i^{\mathrm{ICL}}\bigr).
    \label{eq:iw_alloc}
\end{equation}
Here, $s_i^{\mathrm{ICL}}$ represents the relative positive contribution of
$c_i$, while $s_i^{\mathrm{IW}}$ increases when the model can recover the
response without context and the segment has limited positive contribution.

The predicted source label is obtained by comparing the two proxy scores:
\begin{equation}
\mathcal{I}_i
=
\begin{cases}
\text{\ac{icl}},
&
\text{if }
s_i^{\mathrm{ICL}}>0
\text{ and }
s_i^{\mathrm{ICL}}\geq s_i^{\mathrm{IW}},
\\[2pt]
\text{\ac{iw}},
&
\text{otherwise}.
\end{cases}
\label{eq:cc_label}
\end{equation}
Equivalently, when
$s_i^{\mathrm{ICL}}+s_i^{\mathrm{IW}}>0$, a segment is labeled \ac{icl} if
\begin{equation}
    \frac{s_i^{\mathrm{ICL}}}
    {s_i^{\mathrm{ICL}}+s_i^{\mathrm{IW}}}
    \geq 0.5.
\end{equation}
If $s_j\leq 0$ for every context unit $c_j$, no unit provides positive
in-context support, and all units are labeled \ac{iw}.

\subsection{AttriBoT (Leave-One-Out)}

\paragraph{Original attribution method.}
AttriBoT approximates \ac{loo} context attribution. For each context unit
$c_i$, it measures the reduction in the response log-probability when $c_i$ is
removed from the full context:
\begin{equation}
\begin{aligned}
s_i
={}&
\log p_{\theta_{\mathrm{ft}}}
\bigl(Y\mid C\oplus X\bigr)
\\
&-
\log p_{\theta_{\mathrm{ft}}}
\bigl(
Y\mid (C\setminus\{c_i\})\oplus X
\bigr).
\end{aligned}
\label{eq:attribot_loo}
\end{equation}
A larger positive value of $s_i$ indicates that removing $c_i$ causes a
larger reduction in the likelihood of the response.

\paragraph{Disentanglement extension.}
We first retain only the positive removal effects by defining
\begin{equation}
    s_i^{+}=\max(s_i,0).
\end{equation}
The positive scores are then normalized into attribution-mass fractions:
\begin{equation}
\rho_i
=
\begin{cases}
\dfrac{s_i^{+}}{\sum_{j=1}^{k}s_j^{+}},
& \text{if } \sum_{j=1}^{k}s_j^{+}>0, \\[8pt]
0,
& \text{otherwise},
\end{cases}
\qquad
i\in\{1,\ldots,k\}.
\label{eq:mass_fraction}
\end{equation}
We compare each fraction with the uniform attribution share $\frac{1}{k}$. A segment
that receives an above-uniform share is treated as providing explicit
in-context support, whereas a segment with a smaller share is treated as
being covered primarily by \ac{iw} knowledge. The predicted source label is
therefore
\begin{equation}
\mathcal{I}_i
=
\begin{cases}
\text{\ac{icl}},
&
\text{if }
\displaystyle
\sum_{j=1}^{k}s_j^{+}>0
\text{ and }
\rho_i\geq \dfrac{1}{k},
\\[7pt]
\text{\ac{iw}},
&
\text{otherwise}.
\end{cases}
\label{eq:attribot_label}
\end{equation}
If all context units have non-positive \ac{loo} scores, then
$\rho_i=0$ for every $i$, and all context units are labeled \ac{iw}.

\subsection{TokenShapley}

\paragraph{Original attribution method.}
TokenShapley computes KNN-Shapley values over hidden-state representations.
For each generated response token $y_t$,
we extract the hidden state $\mathbf{h}_t$ from the final transformer layer and compute its cosine distance to every context token position in the same layer.
A weighted $k$-nearest-neighbor Shapley value is then computed at the token level,
measuring each context token's contribution.
Segment-level scores are obtained by aggregating token-level values:
\begin{equation}
    s_i = \sum_{t \in \mathcal{T}(c_i)} \max(s_t, 0)
\end{equation}
where $\mathcal{T}(c_i)$ is the set of token positions corresponding to unit $c_i$.

\paragraph{Disentanglement extension.}
Source labels follow the same above-uniform rule as AttriBoT.
The positive-clipped,
aggregated Shapley scores are normalized:
\begin{equation}
\mathcal{I}_i =
\begin{cases}
\mathrm{ICL}, &
\text{if }
\displaystyle \frac{s_i}{\sum_{j=1}^{k} s_j}
\geq \frac{1}{k}, \\[6pt]
\mathrm{IW}, &
\text{otherwise},
\end{cases}
\qquad i \in \{1,\ldots,k\}.
\end{equation}
Units receiving disproportionate Shapley mass in hidden-state space are identified as \ac{icl} sources that actively shape the model's generation,
while units with negligible representation influence are classified as \ac{iw}.

\subsection{TracLLM}

\paragraph{Original attribution method.}
TracLLM uses hierarchical binary-tree perturbation with top-$K$ pruning.
Starting from a single root node containing all context units,
the algorithm recursively bisects each node and scores children using one or more scoring functions.
At each level, only the top-$K$ highest-scoring nodes are retained,
and recursion continues until all nodes are singletons.
The final segment score is inherited from its containing node.

Two scoring functions are combined via a weighted maximum ensemble:
\begin{itemize}
    \item \textbf{STC} (Single Text Contribution): $\text{STC}(S) = \log p_{\theta_{\text{ft}}}(Y \mid S \oplus X) - \log p_{\theta_{\text{ft}}}(Y \mid X)$, measuring a node $S \subseteq C$'s standalone contribution relative to the no-context baseline.
    \item \textbf{LOO}: $\text{LOO}(S) = \log p_{\theta_{\text{ft}}}(Y \mid C \oplus X) - \log p_{\theta_{\text{ft}}}(Y \mid (C \setminus S) \oplus X)$, measuring the effect of removing a node from the full context.
\end{itemize}
The final score for each unit is $s_i = \max(w_{\text{loo}} \cdot \text{LOO}_i,\; \text{STC}_i)$,
where $w_{\text{loo}} = 2.0$ is a weight hyperparameter.

\paragraph{Disentanglement extension.}
TracLLM uses the same contrastive interface as ContextCite.
After computing the base TracLLM scores $s_1, \ldots, s_k$, the segment-level
allocations $s^{\mathrm{ICL}}_i$ and $s^{\mathrm{IW}}_i$ and the predicted
source labels $\mathcal{I}_i$ follow
Equations~\eqref{eq:icl_alloc}--\eqref{eq:cc_label}.

% \section{Suppression and memorization}
% \begin{figure*}[!t]
%     \centering
%     \includegraphics[width=1\textwidth]{figures/suppression_vs_memorization_3models.pdf}
%     \caption{
%     Score suppression (base score minus finetuned score)
%     versus memorization confidence (no-context logit gap of the
%     finetuned model) for in-weight segments. 
%     A positive correlation indicates stronger memorization, 
%     predictably suppresses attribution scores.
%     }
% \label{fig:score_supression}
% \end{figure*}

\section{Additional details for the WMDP-Cyber++ dataset creation}
We provide the prompt for each step to use WMDP dataset for creating WMDP-Cyber++ with the GPT-4o model.

\begin{figure}[ht]
\begin{tcolorbox}[
    title=Stage 1: In-weight segment re-ranking (GPT-4o),
    colback=gray!5,
    colframe=black,
    fonttitle=\bfseries,
    boxrule=0.5pt
]
\small
You are an expert at evaluating text relevance for cybersecurity questions. \\
\noindent Given the question below, score each candidate passage on how relevant it is to the question (0--10). A passage scores high if it contains information directly related to concepts, 
techniques, or knowledge needed to answer the question. \\
Return ONLY a JSON array of numeric scores, one per passage, in order. \\
\noindent \textbf{Question:} \\
\emph{\{WMDP-cyber question with choices\}} \\
\noindent \textbf{Passages:} \\
{[1]} \emph{\{Retrieved chunk from \texttt{cyber-retain-corpus}\}} \\
{[2]} \emph{\{...\}} \\
\dots \\
\noindent \textbf{Scores (JSON array):}
\end{tcolorbox}
\end{figure}

\begin{figure}[ht]
\begin{tcolorbox}[
    title=Stage 2: In-context segment filtering (GPT-4o),
    colback=gray!5,
    colframe=black,
    fonttitle=\bfseries,
    boxrule=0.5pt
]
\small
You are an expert at evaluating whether text passages provide supporting evidence for answering a question. \\
\noindent Given the question and correct answer below, 
score each passage on how well it provides supporting evidence or background knowledge that helps reach the correct answer (0--10). \\
A passage scores high if it contains relevant facts, context, or technical details that support answering the question correctly. \\
A passage scores 0 if it is completely irrelevant. \\
Return ONLY a JSON array of numeric scores, one per passage, in order. \\
\noindent \textbf{Question:} \\
\emph{\{WMDP-cyber question with choices\}} \\
\textbf{Correct answer:} \emph{\{letter\}. \{answer text\}} \\
\noindent \textbf{Passages:} \\
{[1]} \emph{\{Retrieved chunk from \texttt{cyber-forget-corpus}\}} \\
{[2]} \emph{\{...\}} \\
\dots \\
\noindent \textbf{Scores (JSON array):}
\end{tcolorbox}
\end{figure}

\begin{figure}[ht]
\begin{tcolorbox}[
    title=Stage 3: Context smoothing (GPT-4o),
    colback=gray!5,
    colframe=black,
    fonttitle=\bfseries,
    boxrule=0.5pt
]
\small
You are a technical writer who creates coherent reference documents from multiple source passages. \\
\noindent Rewrite the following passages into a single coherent reference document that could serve as context for answering a technical question. \\
\textbf{Rules:} \\
1. Preserve ALL factual content from every passage --- do not drop information. \\
2. Keep the [S1], [S2], etc.\ markers at the beginning of each passage's content so we know which source each part came from. \\
3. Add brief transitional phrases between sections so the document reads naturally and coherently. \\
4. Do NOT add new facts, opinions, or information not in the originals. \\
5. Keep the technical level and terminology of the originals. \\
6. Do NOT mention or reference the question in the document. \\
\noindent \textbf{Question} (for context only, do NOT include in output): \\
\emph{\{WMDP-cyber question\}} \\
\noindent \textbf{Passages:} \\
{[S1]} \emph{\{ICL or IW segment\}} \\
{[S2]} \emph{\{...\}} \\
\dots \\
\noindent \textbf{Rewritten document:}
\end{tcolorbox}
\end{figure}

\end{document}